\UseRawInputEncoding
\documentclass[lettersize,journal]{IEEEtran}
\usepackage{amsmath,amsfonts}
\usepackage{algorithmic}
\usepackage{algorithm}
\usepackage{array}
\usepackage[caption=false,font=normalsize,labelfont=sf,textfont=sf]{subfig}
\usepackage{textcomp}
\usepackage{stfloats}
\usepackage{url}
\usepackage{verbatim}
\usepackage{graphicx}
\usepackage{xcolor}
\usepackage{cite}
\usepackage{diagbox}
\usepackage{hyperref}
\usepackage{cite}
\usepackage{float}

\usepackage{lipsum}
\usepackage{graphicx}
\ifCLASSOPTIONcompsoc
    \usepackage[caption=false, font=normalsize, labelfont=sf, textfont=sf]{subfig}
\else

\begin{document}

\title{A low complexity contextual stacked ensemble-learning approach for pedestrian intent prediction}

 \author{Chia-Yen Chiang, Yasmin Fathy, Gregory Slabaugh, Mona Jaber
\thanks{C. Chiang, G. Slabaugh and M. Jaber are with the School
 of Electronic Engineering \& Computer Science, Queen Mary University of London: \{c.chiang, g.slabaugh, m.jaber\}@qmul.ac.uk. Y. Fathy is with the Department of Engineering, University of Cambridge {yafa2@cantab.ac.uk}}
\vspace{-0.55cm} }

\maketitle

\begin{abstract}
{Walking, as a form of active travel, is a fundamental element in promoting sustainable transport. Accurately predicting pedestrian crossing intention is vital in preventing collisions, especially with the rise of autonomous vehicles and the Internet-of-Vehicles (IoV). Current research relies on computer vision and machine learning to predict near-misses, often requiring high computation power to yield reliable results. In contrast, our work presents a low-complexity ensemble-learning method that utilizes contextual data to predict pedestrian intent for crossing. The pedestrian is first detected, and their image is compressed using skeleton-ization. Next, we incorporate contextual information into a stacked ensemble-learning model. Our experiments conducted on different datasets show that our approach achieves a similar pedestrian intent prediction performance to the state-of-the-art approaches, with a $99.97\%$ reduction in computational complexity. Our source code and trained models will be released upon paper acceptance.}
\end{abstract}


\section{Introduction}\label{sec:intro}
\IEEEPARstart{A}{ctive} travel is critical in the pathway to sustainable and effective urban transportation systems. Active travel is a transportation mode in which people travel from one place to another by being physically active, such as walking or cycling. However, the growing concerns about the safety of vulnerable road users (VRUs), such as pedestrians, have slowed the uptake of active travel. 
According to the World Health {Organization} data, $1.19$ million people die each year as a result of road traffic crashes, and more than half are VRUs\footnote{https://www.who.int/news-room/fact-sheets/detail/road-traffic-injuries}.

The future vision of urban mobility consists of various modes of transport sharing the open urban space and road network, including VRUs, connected and autonomous vehicles as part of the Internet of Vehicles (IoV)~\cite{taslimasa2023security}.  
Human drivers have `common sense' that allows them to `read' a scene, predict the behavior of VRUs, and communicate with them through eye contact to indicate safety to crossing~\cite{Eye}. Can we train autonomous vehicles to develop a similar `common sense' that predicts the behavior of VRUs? Pedestrian intent prediction (PIP) has gained attention recently, particularly due to advancements in autonomous vehicles within the intelligent transportation systems framework.

{
PIP methods often rely on technologies used in connected and autonomous vehicles (CAVs) and utilize a variety of sensors, including radar, LIDAR, optical cameras, and ultrasonic sensors. These sensors leverage various machine learning (ML) techniques to automate decision-making process that helps avoid collisions~\cite{CAVsensors}. Recent advances in computer vision have greatly improved the ability to identify pedestrians and PIPs accurately. However, the data collected by cameras is often complex and requires high computation power for processing. Furthermore, this data may contain personal information about pedestrians, which could pose a risk if misused while being irrelevant to addressing the PIP problem.}
 
 {
There are two primary approaches for computer vision (CV)-enabled PIP. The first approach relies on cameras installed on the roads, which require reliable broadband connectivity to a cloud server that offers high computational power~\cite{Zhang20}. However, this cloud processing can raise privacy concerns, mainly when personal information (e.g., people's faces) is transmitted over the network for analysis. Some research addresses this issue using edge computing to offload pedestrian-related prediction tasks and improve reliability. The second approach involves collecting and processing data from cameras and other sensory devices mounted on vehicles. This approach is essential for developing advanced driver-assistance systems (ADAS) and IoV. Integrating ADAS with IoV offers opportunities to improve safety, decision-making, and efficiency in driving.}

{
 With the emergence of 5G and next-generation networks, broadband connectivity for moving vehicles is possible. However, this connectivity tends to be less reliable and more costly compared to the connectivity of fixed cameras~\cite{toktam}. 
 Moreover, the robustness of the sensors and of the data fusion and artificial intelligence (AI) algorithms varies significantly among different vendors. As a result, centralized learning is often preferred to ensure the reliability of AI-based decisions. Therefore, data communication to the central server remains predominant in this approach, but it faces challenges due to the large volumes of data that need to be transferred~\cite{Cadena22}. }

{
A growing concern shared by both approaches is their significant overhead resulting in computationally complex AI models, which contribute to greenhouse gas emissions. This issue is emphasized in a study where computational researchers are encouraged to monitor and report the carbon footprint of their work to promote a more sustainable future~\cite{nature}. The relation between computational complexity and carbon footprint is highlighted in~\cite{EcologicalInformatics}, where the study acknowledges the essential role of AI in facilitating sustainable outcomes while advocating for the development of efficient and low-complexity AI models.}

{
This work proposes a new method called Contextual Stacked Ensemble-learning (CSE) for solving PIP based on data captured by front car-mounted cameras. Similar to previous studies \cite{yang2024faster}, \cite{PedGraphPlus}, \cite{Cadena22}, we employ a process known as skeleton-ization, which compresses the image of a pedestrian into 17 keypoints. This approach reduces the volume of camera-based data by 4,336 times (see Section~\ref{sec:skltn}), while also addressing privacy concerns. As a result, it becomes feasible to transmit the skeletonized data to a cloud server for further processing and to perform deep intent prediction.}

{
In recent research,  Graph Neural Networks are proposed to model the change of human keypoints in time, which helps represent pedestrian crossing behavior\cite{yang2024faster} \cite{PedGraphPlus} \cite{Cadena22}. Other studies \cite{ahmed2023multi} employ Long Short-Term Memory (LSTM) to model the relationship between pedestrian locations and crossing behavior. A transformer network has also been introduced in~\cite{zhang2023trep} for PIP to capture the AI uncertainty in scene complexities. This research has advanced the PIP performance but such methods often require high computational complexity and have a significant impact on greenhouse gas emissions. }

{
To this end, the main objectives of our proposed method are to ensure robust and reliable PIP with low computational complexity and data compression to enhance privacy. The robust and reliable PIP model is achieved by leveraging contextual information, such as categorical data regarding the scene and pedestrian trajectory, through stacked ensemble learning. The proposed CSE approach achieves comparable PIP performance to state-of-the-solutions while significantly reducing complexity, achieving ($99.99\%$ reduction in FLOPS and $99.97\%$ less trainable parameters).
Moreover, our approach won the first prize award of the IEEE Intelligent Transportation Systems Society (ITSS) Student Competition in Pedestrian Behavior Prediction \cite{psicompetition}.}

{
\section{Problem formulation}
In this work, the PIP problem is formulated as predicting the behavior of a pedestrian between \textit{Crossing} or \textit{Not-Crossing} based on video clips taken from a camera mounted at the front of a vehicle.} Each video clip, denoted as $\mathbb{S}_v$, for $v \in \{1, \dots, V\}$ ($V$ is the total number of clips in the dataset), contains a consecutive set of $M_v$ frames denoted as $F_i$, where $i$ is the index of frames $i\in \{1, \dots, M_v\}$. Each clip $\mathbb{S}_v$ includes a maximum of $P_v$ bounding boxes (\textit{bbx}), and each \textit{bbx} bounds a single pedestrian with an index $j$ and may include other pedestrians in the background (i.e., not labeled). We note that video clips differ in length $M_v$ and include a different number $P_v$ of captured pedestrians. It follows that each annotated clip includes the labeled behavior of $P_v$ pedestrians with a unique index $j \in \{1, \dots, P_v\}$. {Given the one-to-one mapping between \textit{bbx} and labeled pedestrians,} the same index $j$ is used to reference the \textit{bbx} that bounds a pedestrian with index $j$.

{A sliding windows of width $F_{max}=32$ frames is used to span each full video clip to generate segments with index  $s \in \{1, \dots, M_v/F_{max}\}$ that describe the movement of pedestrian with index $j$ in a video clip with index $v$.}

In each video clip $\mathbb{S}_v$, a pedestrian with index $j \in \{1, \dots, P_v\}$ (and the \textit{bbx} that frames them) is annotated with a label $y$; $y=1$ \textit{Crossing} if pedestrian intends to cross a street at any point in $\mathbb{S}_v$ or $y=0$ for (\textit{Not-Crossing} or \textit{Irrelevant}, i.e., without a behavioral label), otherwise. 

{The PIP problem can thus be formulated as a classification predictive modeling with the objective of predicting the label $\hat{y}_{v,i}$ associated with each segment $p_{v,i,s}$ of a pedestrian bounded by \textit{bbx} with index $j$, in sliding window $s$ of video clip $\mathbb{S}_v$. It follows that this problem is concerned with maximizing the number of correctly predicted labels ${\hat{y}_{v,j,s}}$ for all segments $p_{v,j,s} \in \mathbb{P}$ and $\forall v \in \{1, \dots, V\}, \forall j \in \{1, \dots, P_v\}$, and $\forall s \in \{1, \dots, M_v/F_{max}\}$, as formulated here:}
\begin{equation}
\max ~~~~~ \sum_{v=1}^{V}\sum_{j=1}^{P_v}\sum_{s=1}^{\frac{M}{F_{max}}}  (\hat{y}_{v,j,s}=y_{v,j,s})\label{eq:PIP}
\end{equation}

{Towards solving the problem in~\eqref{eq:PIP}, we need to first identify the features that describe the segments in $\mathbb{P}$. To this end, the data in each video clip are analyzed from two perspectives. The first examines the movement of target pedestrians by studying the corresponding \textit{bbx} and its changing relative position to the frame. }Two data streams are extracted from this perspective: keypoints that represent the pose of the pedestrian (see Section~\ref{sec:skltn}) and the pedestrian's trajectory relative to the frame (see Section~\ref{sec:track}). The other perspective captures the contextual information from two angles: global context and local context. The former is common to all pedestrians in a given frame and describes the speed of the ego-vehicle, the status of the traffic light it faces, and the road type it drives on. The latter is specific to each pedestrian and describes their relative local context such as standing at an intersection (see Section~\ref{sec:frame}). The goal of this work is therefore to first generate these features that will then be used to classify segments in $\mathbb{P}$ as positive (\textit{Crossing}) or negative (\textit{Not-Crossing} and \textit{Irrelevant}) case. 

\section{Methodology}\label{sec:meth}
\begin{figure*}
    \centering
\includegraphics[width=18cm,height=13cm,keepaspectratio]{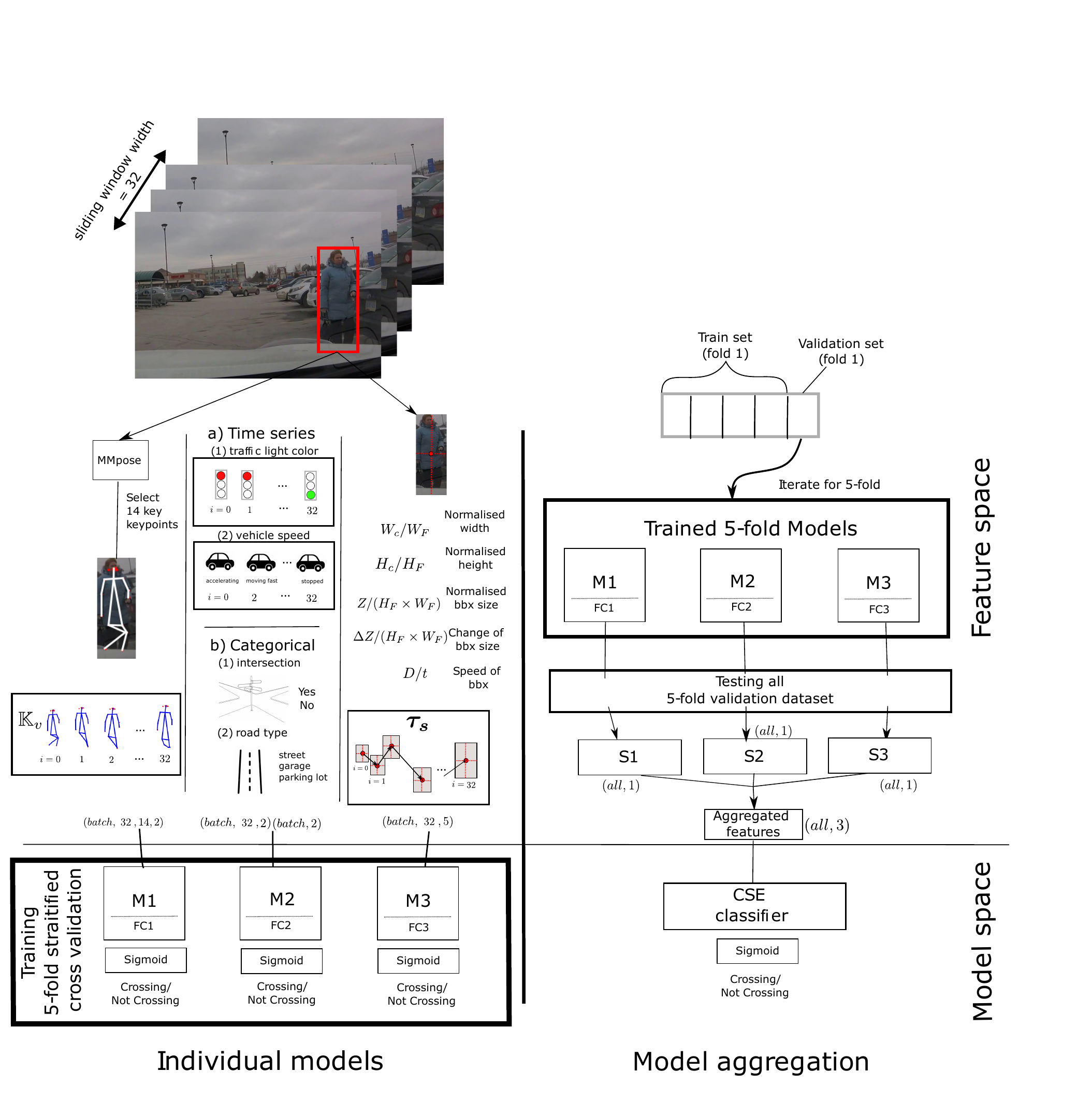}
    \caption{Our proposed CSE framework is composed of a stacked ensemble classifier and three models for PIP classification. (Top-left) Data preprocessing on video frames shows how three representations are extracted from the raw video. (Bottom-left) M1 uses skeleton-ization for data compression followed by a Graph Neural Network. {M2 uses contextual data and a Gated Recurrent Unit network.} M3 extracts the relative location of the pedestrian and uses a one-dimensional convolutional neural network. (Top-right and bottom-right) The aggregated features stack the output of the last fully connected (FC) layer, i.e., S1, S2, S3, from the three models by aggregating testing results from 5-fold validation of their corresponding model, i.e., M1, M2, M3. The system then produces prediction results from the contextual stacked ensemble classifier.}
    \label{fig:stack}
\end{figure*}

{Different methods have been proposed to address PIP in~\cite{pedgraph02, Cadena22}\cite{GRUpaper, GRUpaper2} however, these methods suffer from high computational complexity. 
Therefore, in this work, we} propose {Contextual Stacked Ensemble} (CSE), a novel ensemble learning system that builds on three individual models which we introduce in this section. The first model, M1, consists of a graph neural network (GNN) model for classifying skeleton-ized data, {$\mathbb{K}_v$} as \textit{Crossing}/\textit{Not-Crossing}. {The second model, M2, employs stacked gated recurrent network units (GRUs) for solving PIP using time-series and categorical data from JAAD's vehicle and traffic annotations}. The third model, M3, employs a one-dimensional convolutional neural network (1D-CNN) for solving PIP based on a pedestrian's trajectory, {$\boldsymbol{\tau_v}$}. The CSE combines these two or three models as shown in Figure~\ref{fig:meta} (right). { Each of models M1, M2, M3, and CSE model are structured to represent the PIP of a target pedestrian during a sliding window of $F_{max}$ frames.} {For comparison,} two models are also implemented using a similar structure. The entire ensemble learning system is thus structured as shown in Figure~\ref{fig:stack}.

\subsection{M1: Graph Neural Network}
{The GNN model (M1) is used to classify a sequence of graphs {where vertices are represented by the $K_{max}$ keypoints representing a target pedestrian with index $j \in \{1, \dots, P_v\}$ bounded by a \textit{bbx} in the sliding window with index $s \in \{1, \dots, M_v/F_{max}\}$ of clip with index $v$ and maximum number of indexed pedestrians $P_v$. For simplicity, in the description of M1, we drop the indices related to the video clip and to the pedestrian and focus on the index that tracks the movement of the target pedestrian, i.e., the index of the frame. For a given target pedestrian bounded by a \textit{bbx}, each vertex (i.e., keypoint) in each frame $F_i$ within the sliding window is depicted by a matrix $\mathcal{K}_{i}$, which is obtained through the skeleton-ization process. In this context, $k_\kappa=\{w_\kappa,h_\kappa\}$ denotes the relative position of a keypoint with index $\kappa$, and $\mathbb{K}_v$ (see Figure~\ref{fig:stack}) represents the sequence of matrices $\mathcal{K}_{i}$ throughout a video clip $\mathbb{S}_v$.} 

\begin{eqnarray}
&\mathcal{K}_{i}=[k_1, \dots,k_\kappa], \kappa = \{1,\dots K_{max}\}\\
&k_\kappa=\{w_\kappa,h_\kappa\} \mid  0<w_\kappa<W ; 0<h_\kappa<H\\
&\mathbb{K}_v=\{ \mathcal{K}_{i} : 1\leq i \leq F_{max}\}
\end{eqnarray}%
These keypoints (vertices) are connected with a fixed adjacency matrix $\mathcal{A}$ defined to represent the connectivity between any two different keypoints $k_o$ and $k_q$ in a skeleton, such that $A(o,q)=1$ when these are connected (e.g. head and neck)~\cite{pedgraph02}.
\begin{equation}
  \begin{aligned}
    \mathcal{A}=[\{A(1,1), \dots, A(1,K_{max})\}, \dots, \\
    \{A(\kappa,1),\dots, A(\kappa,K_{max})\}, \dots, \\
    \{A(K_{max},1), \dots, A(K_{max},K_{max})\} ]\
 \end{aligned}
\end{equation}}

\begin{figure}[h]
    \centering
\includegraphics[width=0.4\textwidth]{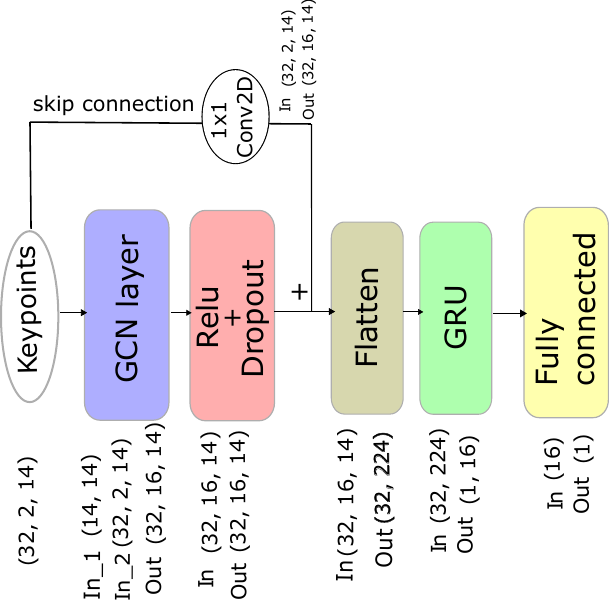}
    \caption{\textbf{M1}: The figure shows the model structure with input and output dimensions in each model block. {Note that the shape of Keypoints at the input of M1 is determined by $F_{max}=32$, 2D coordinates, and $K_{max}=14$. }}
    \label{fig:graph_structure}
\end{figure}

The model consists of a Convolutional Graph Neural Network layer (GCN layer), a Relu+Dropout, a skip connection, a Flatten and a GRU, as shown in Figure~\ref{fig:graph_structure}.
The input keypoints are first applied matrix multiplication with {a} normalized adjacency matrix. The matrix product involves summing up keypoint features from adjacent nodes. The summed-up features of each node are embedded into a convolutional operation in the GCN layer for further feature extraction. To prevent shrinking feature maps, the convolutionaloperation must use {``same"} padding. A Relu and Dropout {(0.5)} layers {are then} used to encourage model generalization.   

{The skip connection operation adds input keypoints to the dropout output. 
By passing coming layers and preserving early layer information, the model can keep both low and high level representations, thus improves feature learning. A 1x1 convolution on the skip connection branch adjusts the channel dimension to enable the addition to the features output by the GCN layer.

Finally, after flattening the merged features from Dropout and skip connection, a GRU layer is used to model the evolution of keypoint features in {the} time domain. The hidden size is set to $16$. The last hidden output is then fed to a fully-connected layer for \textit{Crossing}/\textit{Not-Crossing} classification.}

\subsection*{M2: Stacked Gated Recurrent Unit Network}

\begin{figure}[h]
    \centering
\includegraphics[width=0.32\textwidth]{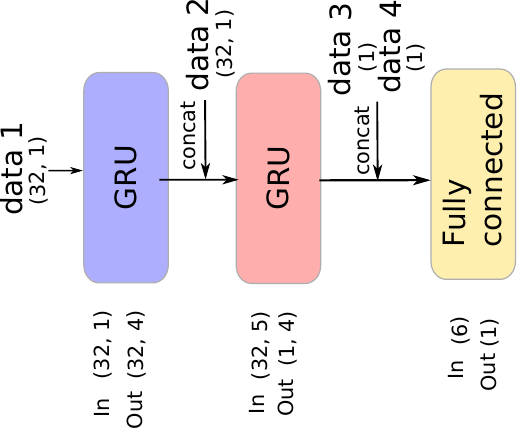}
    \caption{ \textbf{M2}: a stacked GRUs is used to model contextual information that can affect the crossing decision. {Note: data 1 is the vehicle speed, data 2 is the traffic signal status, data 3 is the pedestrian-intersection status, and data 4 is the type of road. The shape of each input data corresponds to the size of the sliding window $F_{max}$.}} \label{fig:M2}
\end{figure}

Inspired by the stacked GRU in \cite{rasouli2020pedestrian}, we apply the stacked GRU in modelling time-series and categorical contextual information from the vehicle and traffic background as shown in Figure~\ref{fig:M2}. The stacked GRUs model (M2) iterates each time-series data and feeds it to a GRU with only one layer to produce all hidden outputs from frame $F_{1}$ to $F_{max}$. Before feeding all hidden outputs to the next GRU, a next time-series data is merged with the hidden outputs. The output of the last GRU is the last hidden output instead of a full sequence of hidden outputs. 

Specifically, the categorical data used for contextual information in M2 consists of the following data (Figure~\ref{fig:M2}): 
\begin{itemize}
    \item Data 1 (time-series data over the duration of $F_{max}$ frames) - vehicle speed \{\textit{Stopped, Slow, Fast, Accelerating, Decelerating}\}: A pedestrian is less likely to cross if they notice an upcoming vehicle that is speeding.
    \item Data 2 (time-series data over the duration of $F_{max}$ frames) - Traffic signal status \{\textit{Red, Green}\}: A pedestrian is less likely to cross the street if the traffic signal controlling the passing vehicles is green; in which case the pedestrian signal would be red.
    \item Data 3 (categorical with one value for each \textit{bbx})- Is the pedestrian standing at an intersection? \{\textit{Yes, No}\}: A pedestrian waiting at an intersection is likely to cross the street.
    \item Data 4 (categorical with one value for each video)- Type of road \{\textit{Garage, Parking lot, Street}\}: It is more likely that a pedestrian seen on the street might cross than a pedestrian standing/walking in a parking lot.
\end{itemize}

Our proposed approach is different from \cite{rasouli2020pedestrian} in the following. The last hidden output of stacked GRUs in the proposed M2 is concatenated with categorical data. The combined features then are projected to a fully connected layer before \textit{Crossing}/\textit{Not-Crossing} classification (see Figure~\ref{fig:M2}).

\subsection{M3: 1D convolutional Neural Network}
\begin{figure}[h]\label{fig:1dcnn_structure}
    \centering
\includegraphics[width=0.32\textwidth]{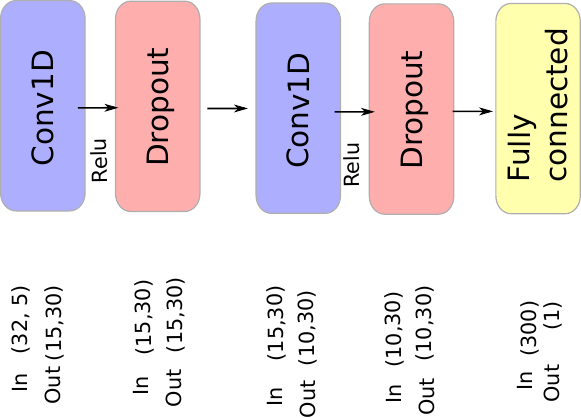}
    \caption{ \textbf{M3}: A 1D-CNN could be used to predict future trajectory given historical data. The five inputs of the trajectory are introduced in Section~\ref{sec:track}. { Note that the input dimension of the first Conv1D is ($F_{max}$, five trajectory attributes) = $(32, 5)$. They are normalized bounding box height, width and size, the change of size and the speed of the box.} }
\end{figure}

{For} modelling the relationship between pedestrian trajectory $\boldsymbol{\tau_v}$ and PIP decision, a shallow 1D-CNN (M3) is proposed {(see Section~\ref{fig:1dcnn_structure})}. 1D-CNN is often used to predict a future trajectory given historical data\cite{zamboni2022pedestrian,nikhil2018convolutional} while~\cite{abu2018will02} uses both RNN and CNN to predict future classes of an encoded trajectory. Trajectories possess not only pedestrian speed, direction, and pedestrian depth information but are also strongly related to PIP decisions. We, therefore, adopt a 1D-CNN for M3 to solve PIP based on past trajectory as contextual information. Our proposed 1D-CNN model contains only three layers: a convolutional layer for extracting 1D time-series features, a dropout layer for preventing overfitting, and a fully connected layer for PIP classification, as shown in Figure \ref{fig:1dcnn_structure}. 

\subsection{CSE: Contextual Stacked Ensemble}
\begin{figure*}[!htbp] 
    \centering
\includegraphics[width=0.65\textwidth]{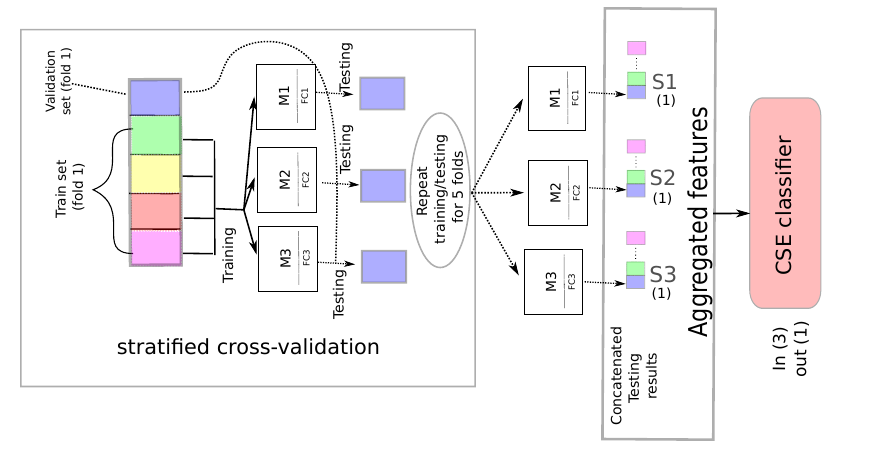}
    \caption{ \textbf{CSE}: Model structure of Contextual Stacked Ensemble learning. {In stratified cross-validation, we first train models with five-fold training datasets. An example is shown in Fold 1, where three models are trained on four blocks (see bottom-to-top 4 blocks marked train set Fold 1 above). In this case, the validation block (top purple) is used to test each of the three models. This process is repeated with the iteration of folds; for instance, the next fold would use the green block for validation and train using the remaining blocks. The testing results of all five folds are concatenated to form S1, S2, and S3 (or a subet) as an input of CSE classifier. }   \label{fig:meta}}
\end{figure*}

Similar to the approach used in~\cite{zhou2021multimodal} for multimodal feature fusion, we propose a stack-ensemble learning method to fuse features from two or three models defined above.
Each of M1, M2, and M3 is trained by a five-fold stratified cross-validation
with a fixed sampling seed 1. The variant of cross-validation ensures each fold preserves the same class ratio. The last fully connected layer of each of the three models M1, M2, and M3 produces an output S1, S2, or S3, respectively (as shown in Figures~\ref{fig:meta} and~\ref{fig:stack}) when the test input is a five-fold validation dataset.  
We then concatenated S1, S2, S3 (or a subset) to form the aggregated features that are then fed into a CSE classifier to boost overall prediction. The classifier is then followed by the sigmoid and {argmax} functions to generate the final \textit{Crossing}/\textit{Not-Crossing} classification results (see CSE model structure in Figure \ref{fig:meta}).

\section{Experiments}\label{sec:res}
{We conducted the experiments of our proposed method on two different labeled versions of the Joint Attention in Autonomous Driving (JAAD) dataset: JAADbeh\cite{rasouli2017they} and JAADall, which includes additional annotations of pedestrians as described in Section~\ref{sec:data}. The rest of this section discusses data engineering, where a sequence of $F_{max}$ frames (defined by the sliding window) is processed, and three different groups of features are extracted. Firstly, the keypoint representation of labeled pedestrians bounded by \textit{bbx} is discussed in Section~\ref{sec:skltn}. Secondly, extracting contextual information around pedestrians, including time-series and categorical data, is described in Section~\ref{sec:frame}).} The third group of features describes the historical trajectory of the pedestrian, which is extracted by analyzing the changing position of the \textit{bbx} with respect to the frame (Section~\ref{sec:track}).
{
\subsection{Dataset description}\label{sec:data}
In this work, we use the JAAD Dataset, which is richly annotated and contains $V=346$ video clips (each is around five to ten seconds long) {extracted from} over $240$ hours of driving footage from dash cameras positioned at the front of the car. The dataset provides frame-wise, distinct labels including pedestrian reaction toward drivers, pedestrians' demographic attributes, and traffic conditions, which are not found in other datasets, such as the Multiple Object Tracking (MOT) Challenge \footnote{\url{https://motchallenge.net}}. There are two versions of the same dataset, JAADbeh and JAADall, where JAADbeh is a subset of JAADall, and the latter includes additional annotations of pedestrians in the background.~\footnote{JAADall dataset incorporates all non-labeled pedestrians in JAADbeh by manually annotating these to \textit{Not-Crossing} class. (see details in Section~\ref{subsec:datapreprocess})}. }

\subsection{Feature Engineering}
\label{subsec:datapreprocess}
\subsubsection{Skeleton-ization}\label{sec:skltn}
The first data representation aims to capture the pose of a pedestrian by examining the essential keypoints of the human body following a skeleton-ization approach. The reasoning behind this representation is multifaceted.
First, it avoids the capturing of personal information (e.g., face, hair, clothing, etc.) while retaining the essential information for pose analysis and, therefore, action prediction. {In this work, we first apply the MMpose\footnote{https://github.com/open-mmlab/mmpose/tree/main} algorithm which returns $17$ key points. Next, we combine the keypoints representing the left eye, right eye, and nose into a single keypoint that represents the head. Similalry, we replace the keypoints representaing the left and right shoulders with a single point in the middle that represents the thorax. It follows that our M1 model is based on $14$ keypoints instead of $17$.}
\begin{figure}[!htbp] \label{fig:keypoints}
    \centering
\includegraphics[width=0.3\textwidth]{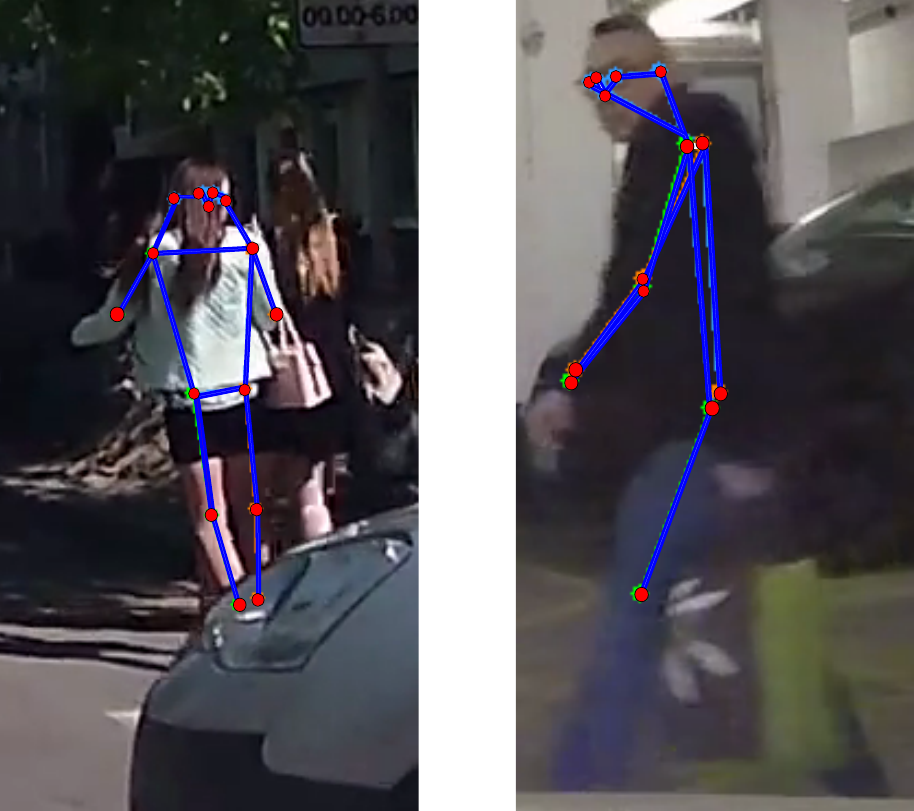}
    \caption{The MMpose returns $17$ keypoints for each \textit{bbx}. (Left) \textit{Not-Crossing} (right) \textit{Crossing }{. Note: In this work, the $17$ keypoints are further reduced to $14$ (see Section \ref{sec:skltn}).} }
\end{figure}

\begin{figure*}[!htbp] 
    \centering
\includegraphics[width=0.7\textwidth]{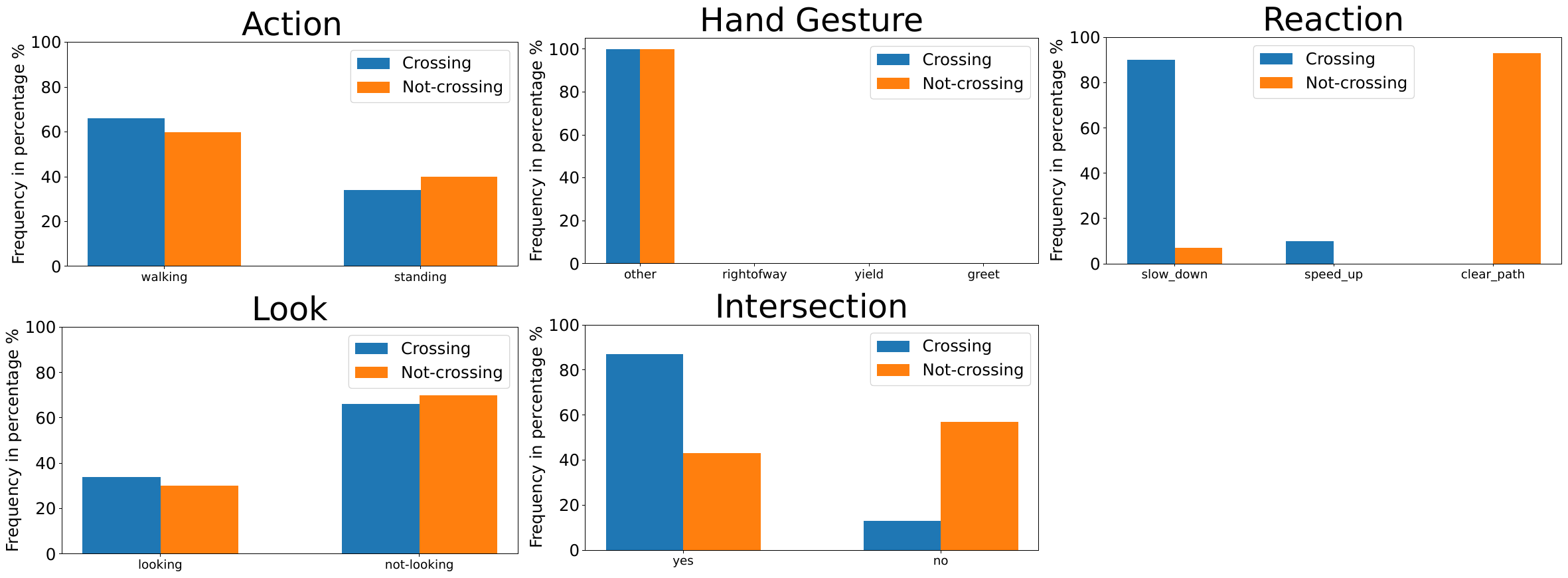}
    \caption{Pedestrian Behavior comparison between \textit{Crossing} and \textit{Not-Crossing} before crossing action. Intersection and Reaction have the most distinguishable patterns in subcategories for both classes.}\label{fig:behavior}
\end{figure*}

\begin{figure*}[!htbp] 
    \centering
\includegraphics[width=0.7\textwidth]{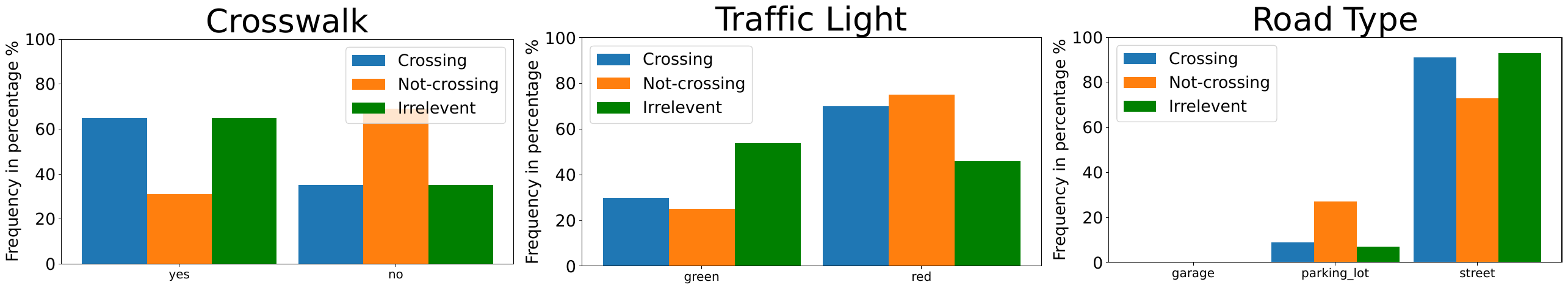}
    \caption{Traffic Annotation comparison between \textit{Crossing}, \textit{Not-Crossing} and \textit{Irrelevant}. It is clear that \textit{Irrelevant} has similar patterns with \textit{Crossing} in Pedestrian Crossing, and Road type. In addition, Traffic Light can't differentiate \textit{Crossing} and \textit{Not-crossing} but Pedestrian Crossing and Road Type can.}\label{fig:traffic}
\end{figure*}

\begin{figure}[!htbp] 
    \centering
\includegraphics[width=0.25\textwidth]{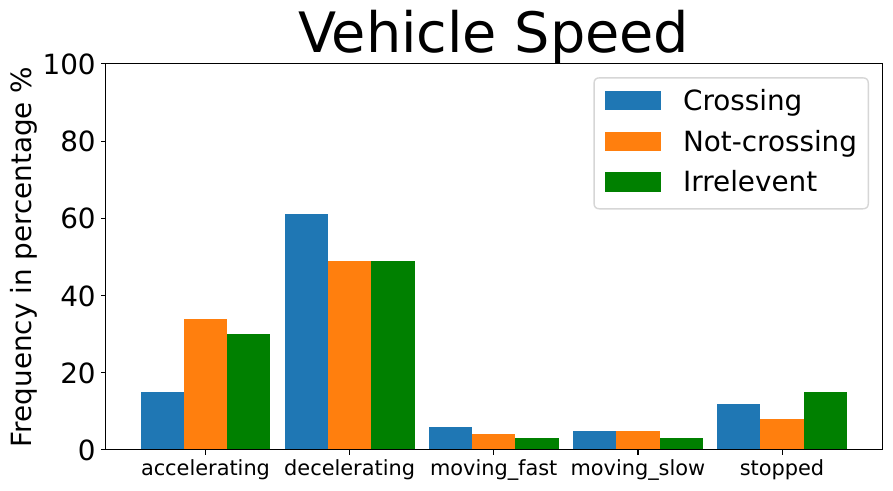}
    \caption{Vehicle Speed comparison between \textit{Crossing}, \textit{Not-Crossing} and \textit{Irrelevant} before crossing action. Vehicle speed seems to differentiate \textit{Crossing} with the group of \textit{Not-crossing} and \textit{Irrelevant}.}\label{fig:vehspeed}
\end{figure}
\subsubsection{Contextual data}
\label{sec:frame}
{
There is a plethora of information available to connected vehicles (see Section~\ref{sec:intro}) that can be inferred from the sensors or location-based services.

{In this section, we explore how well contextual features particular to JAAD can represent pedestrians' intentions. There are three categories of attributes that can be obtained: 
\begin{itemize}
    \item Pedestrian {Behavior} shown in Figure~\ref{fig:behavior}. It includes \textbf{Action}=\{\textit{Walking, Standing}\},  \textbf{Hand Gesture}=\{\textit{Greet, Yield, Rightofway, other}\}, \textbf{Reaction}=\{\textit{Slow-down, Speed-up, Clear-path}\}, Look=\{\textit{Yes, No}\}, and \textbf{Intersection}=\{\textit{Yes, No}\}. All except \textbf{Intersection} are time-series data.
    \item Traffic Background shown in Figure~\ref{fig:traffic} and includes \textbf{Crosswalk} (or Pedestrian Crossing)=\{\textit{Yes, No}\}, \textbf{Traffic Light}=\{\textit{Green, Red}\}, and \textbf{Road Type}=\{\textit{Garage, Parking lot, Street}\}). All except \textbf{Road Type} are time-series data.
    \item Vehicle speed before pedestrian's crossing action shown in Figure~\ref{fig:vehspeed}
\end{itemize}

Pedestrians that are not labeled as \textit{Crossing} or \textit{Not-Crossing} have been marked as \textit{Irrelevant} in Figures~\ref{fig:behavior},~\ref{fig:traffic}, and~\ref{fig:vehspeed}. {Overall, some attributes play a vital role in predicting labels that signify a pedestrian's intention to either \textit{Crossing} or \textit{Not-Crossing}. In the following section, we explore how these features contribute to label prediction.}

It may be counter-intuitive at first that some of the \textbf{Pedestrian Behavior} attributes shown in Figure~\ref{fig:behavior} do not have a distinct pattern for \textit{Crossing} or \textit{Not-Crossing}. {This is particularly evident in the case of pedestrians' intentions, where those who are looking and those who are walking show a near-equal likelihood to \textit{Crossing} or \textit{Not-Crossing}}. However, careful analysis indicates that pedestrians who \textit{stand}are more likely to intend to cross, and those who \textit{look} may be standing-by. \textit{Reaction} shows that \textit{Crossing} pedestrians are more likely to slow down while \textit{Not-Crossing} are more likely to clear path. \textit{Intersection} shows that \textit{Crossing} pedestrians are likely to stand on the intersection. 

{While examining the \textbf{Traffic Annotation} attributes, it seems \textit{road type} and \textit{crosswalk} are similar for both the \textit{Crossing} and \textit{Irrelevant} labels, which indicates that they carry weak representation of pedestrian intent.} Moreover, pedestrians labeled as \textit{Irrelevant} are more likely to be seen in the traffic scene when the traffic light is green than the other two classes. Following a careful investigation of such occurrences, we noted in most cases, pedestrians are not in the same lane as the ego-vehicle. Instead, they could be on the zebra crossing parallel to the ego-vehicle's lane or on the sidewalks.

Further examining the \textbf{Traffic Annotation} attributes in Figure~\ref{fig:traffic}, we can see \textit{Crossing} pedestrians are more likely to appear in the traffic scene with \textit{Crosswalk} (Pedestrian Crossing) and on the \textit{Street} rather than in \textit{Garage}, comparing to \textit{Not-Crossing}. 
As for \textbf{Vehicle Speed} attribute in Figure~\ref{fig:vehspeed}, it is intuitive to note that vehicles \textit{Decelerate} at the sight of pedestrians irrelevant of what their PIP. However, they are more likely to \textit{Decelerate} in front of \textit{Crossing} pedestrians and likely to \textit{Stop} when facing \textit{Crossing} and \textit{Irrelevant} pedestrians.

\subsubsection{Pedestrian trajectory}\label{sec:track}
The trajectory of the pedestrian within each given sliding window carries contextual information that can be useful in solving the PIP problem. The location of the pedestrian is equated to the relative position of \textit{bbx} with respect to the frame. In this work, we calculate the relative position of the central \textit{bbx} pixel located at $[W_c=W/2,H_c=H/2]$, where $W$ and $H$ are the width and height of the \textit{bbx}. The pedestrian's depth, $Z$, is inferred from the size of the \textit{bbx} where a small value indicates that the \textit{bbx} is far away from the camera and vice-versa. Knowing that the absolute coordinates $(W_c,H_c)$ and the size $Z$ can be biased by different frame sizes, we normalize the values such that the pedestrian normalized location is $(W_c/W_F,H_c/H_F)$ and the normalized size is $Z/(H_F\times W_F)$, where $H_F$ is the height of the frame and $W_F$ is the width. In addition, we calculate the change of normalized size over time $\Delta Z/(H_F\times W_F)$ 
as well as the speed of movement, which was estimated by getting the ratio of the normalized traveled distance over time.

For each sliding window with index $s \in \{1, \dots, M_v/F_{max}\}$, the normalized positioning vector $q_{s,i}=[{W_c}_i/W_F, {H_c}_i/H_F, Z_i/(H_F\times W_F),  \Delta Z/(H_F\times W_F), D/t]$ of each \textit{bbx} containing a pedestrian in \textit{bbx} $F_i$ ($i=[0, \dots, F_{max}$]) is calculated. Thus, the trajectory of a pedestrian in sliding window with index with index $s$ is represented by $\boldsymbol{\tau_s}$. 

\begin{table*}[h!]

\caption{Performance and complexity of proposed models using JAADbeh and JAADall. Note: Values shown for Faster-PCPNet, Global PCPA, and PedGrpah+ are not reproduced; these are copied from~\cite{yang2024faster},\cite{yang2022predicting}, and ~\cite{Cadena22} instead.}\label{tab:AllData}
\centering

\begin{tabular}{|l|lllll|lllll|ll|}
\hline
Dataset                                                & \multicolumn{5}{c|}{\textit{JAADbeh}}                                                                        & \multicolumn{5}{c|}{\textit{JAADall}}                                                                        & \multicolumn{2}{c|}{Complexity}                           \\ \hline
                                                       & \multicolumn{1}{l|}{Acc} & \multicolumn{1}{l|}{AUC} & \multicolumn{1}{l|}{F1} & \multicolumn{1}{l|}{P}  & R  & \multicolumn{1}{l|}{Acc} & \multicolumn{1}{l|}{AUC} & \multicolumn{1}{l|}{F1} & \multicolumn{1}{l|}{P}  & R  & \multicolumn{1}{l|}{Flops(K)}          & Params(K)        \\ \hline
M1                                                     & \multicolumn{1}{l|}{58}  & \multicolumn{1}{l|}{65}  & \multicolumn{1}{l|}{58} & \multicolumn{1}{l|}{61} & 61 & \multicolumn{1}{l|}{81}  & \multicolumn{1}{l|}{85}  & \multicolumn{1}{l|}{62} & \multicolumn{1}{l|}{60} & 78 & \multicolumn{1}{l|}{$1.54\times 10^3$} & 12.5             \\ \hline
M2                                                     & \multicolumn{1}{l|}{66}  & \multicolumn{1}{l|}{76}  & \multicolumn{1}{l|}{66} & \multicolumn{1}{l|}{72} & 70 & \multicolumn{1}{l|}{64}  & \multicolumn{1}{l|}{64}  & \multicolumn{1}{l|}{48} & \multicolumn{1}{l|}{53} & 62 & \multicolumn{1}{l|}{6.92}              & 0.22             \\ \hline
M3                                                     & \multicolumn{1}{l|}{61}  & \multicolumn{1}{l|}{69}  & \multicolumn{1}{l|}{61} & \multicolumn{1}{l|}{65} & 64 & \multicolumn{1}{l|}{68}  & \multicolumn{1}{l|}{79}  & \multicolumn{1}{l|}{52} & \multicolumn{1}{l|}{56} & 72 & \multicolumn{1}{l|}{24.6}              & 0.7              \\ \hline
CSE(M1+M2)                                                  & \multicolumn{1}{l|}{59}  & \multicolumn{1}{l|}{67}  & \multicolumn{1}{l|}{58} & \multicolumn{1}{l|}{62} & 62 & \multicolumn{1}{l|}{82}  & \multicolumn{1}{l|}{84}  & \multicolumn{1}{l|}{63} & \multicolumn{1}{l|}{61} & 78 & \multicolumn{1}{l|}{$1.55\times10^3$}  & 12.72            \\ \hline
CSE(M1+M3)                                                  & \multicolumn{1}{l|}{60}  & \multicolumn{1}{l|}{70}  & \multicolumn{1}{l|}{59} & \multicolumn{1}{l|}{65} & 63 & \multicolumn{1}{l|}{83}  & \multicolumn{1}{l|}{87}  & \multicolumn{1}{l|}{63} & \multicolumn{1}{l|}{61} & 79 & \multicolumn{1}{l|}{$1.56\times10^3$}  & 13.2             \\ \hline
CSE(M2+M3)                                                  & \multicolumn{1}{l|}{65}  & \multicolumn{1}{l|}{72}  & \multicolumn{1}{l|}{65} & \multicolumn{1}{l|}{70} & 69 & \multicolumn{1}{l|}{75}  & \multicolumn{1}{l|}{81}  & \multicolumn{1}{l|}{57} & \multicolumn{1}{l|}{57} & 74 & \multicolumn{1}{l|}{31.52}             & 0.92             \\ \hline \hline 
CSE(M1+M2+M3)                                         & \multicolumn{1}{l|}{60}  & \multicolumn{1}{l|}{70}  & \multicolumn{1}{l|}{60} & \multicolumn{1}{l|}{64} & 63 & \multicolumn{1}{l|}{\underline{88}}  & \multicolumn{1}{l|}{\underline{87}}  & \multicolumn{1}{l|}{ \textbf{68}} & \multicolumn{1}{l|}{\underline{65}} & \underline{78} & \multicolumn{1}{l|}{$\mathbf{1.57\times10^3}$} & \textbf{13.42}            \\ \hline
Global PCPA \cite{yang2022predicting} & \multicolumn{1}{l|}{-}   & \multicolumn{1}{l|}{-}   & \multicolumn{1}{l|}{-}  & \multicolumn{1}{l|}{-}  & -  & \multicolumn{1}{l|}{83}  & \multicolumn{1}{l|}{86}  & \multicolumn{1}{l|}{63} & \multicolumn{1}{l|}{51} & \textbf{82} & \multicolumn{1}{l|}{$154.3\times10^6$} & $60.9\times10^3$ \\ \hline
Pedgraph+ \cite{Cadena22}             & \multicolumn{1}{l|}{\textbf{70}}  & \multicolumn{1}{l|}{\textbf{70}}  & \multicolumn{1}{l|}{\textbf{76}} & \multicolumn{1}{l|}{\textbf{77}} & \textbf{75} & \multicolumn{1}{l|}{86}  & \multicolumn{1}{l|}{\textbf{88}}  & \multicolumn{1}{l|}{\underline{65}} & \multicolumn{1}{l|}{58} & 75 & \multicolumn{1}{l|}{520$\times10^3$}   & 70.3             \\ \hline
Faster-PCPNet \cite{yang2024faster}   & \multicolumn{1}{l|}{-}   & \multicolumn{1}{l|}{-}   & \multicolumn{1}{l|}{-}  & \multicolumn{1}{l|}{-}  & -  & \multicolumn{1}{l|}{\textbf{89}}  & \multicolumn{1}{l|}{77}  & \multicolumn{1}{l|}{\underline{65}} & \multicolumn{1}{l|}{\textbf{73}} & 58 & \multicolumn{1}{l|}{$40\times10^3$}    & 30               \\ \hline

\end{tabular}
\end{table*}
\subsection{Results and Discussion} \label{subsec:analysis}
We address the problem of PIP for each given input sliding window sampled from each video clip $\mathbb{S}_v$. {The predicted label $\hat{y}$ is a binary variable, where $\hat{y}=1$ for \textit{Crossing} if a pedestrian is predicted to intend to cross a street or $\hat{y}=0$ for (\textit{Not-Crossing})}. In preparation for the model training and testing, we split the dataset into a training, validation, and test subsets based on JAAD's suggestion (see \footnote{\label{note1}\url{https://github.com/ykotseruba/JAAD/tree/JAAD_2.0/split_ids/default}}). The JAAD's train and validation were then merged before splitting again for stratified 5-fold cross-validation~\cite{scikit-learn}. For each fold, the method preserves the percentage of \textit{Crossing} and \textit{Not-Crossing} data points instead of random sampling on the positive ($y=1$, \textit{Crossing}) and negative ($y=0$, \textit{Not-Crossing}, {\textit{Irrelevant})} samples pool. This approach ensures that the training performance of each fold is not biased by a random ratio of classes and preserves all information in the samples. {The stratified 5-fold crossing validation was used only in training models on JAADbeh. This method preserves the class ratio and therefore, ensures the training stability, which works well for JAADbeh. However, with an extreme class distribution, such as JAADall, we find that models perform better when classes are balanced. Therefore, we create a customized 5-fold dataset in which each fold has all positive samples from the merged data and randomly selected negative samples (without replacement) from the merged data to match the number of positive samples. Nonetheless, the test results are derived using the test dataset defined in Footnote~\ref{note1} for both JAADbeh and JAADall in order to enable fair comparison with prior art.} {Moreover, it is worth noting that there exists an inconsistent class distribution between the train and test set, especially for JAADall. In JAADall, the imbalance ratio of \textit{Crossing}$:$\textit{Not-Crossing} is $1:1$ in the train set but $7:93$ in the test set. On the other hand, in JAADbeh, it is $56:44$ in train set but $4:6$ in the test set.}

Each of the three data streams described in Section~\ref{subsec:datapreprocess} is used independently to train three different models, M1, M2, and M3, as shown in Figure~\ref{fig:stack}. The CSE approach shown in~\ref{fig:meta} is applied to each combination of two models (2-model-based CSE) and to the three models (3-model-based CSE). 
We first present the results of each independent model in addition to the different CSE options in Table~\ref{tab:AllData} and compare the performance and complexity of these to the state-of-the-art models. 

{{Our results are compared to the state of the art in both performance and complexity including~\cite{yang2022predicting}, \cite{Cadena22} and \cite{yang2024faster} which were tested using JAADall. We also compared our models with \cite{yang2024faster} using JAADbeh.} As seen in Table~\ref{tab:AllData}, our proposed CSE using M1, M2, and M3 delivers the best F1 score, outperforming the next best models, Pedgrapgh+~\cite{Cadena22} and Faster PCPNet~\cite{yang2024faster}, by $3\%$ at a much lower computational complexity. Our model requires $331$ times less FLOPS than Pedgraph+ and $10,000$ less than Global PCPA; moreover, our model consists of $5.2$ times less trainable parameters than Pedgraph+ and $4500$ times less than Global PCPA. Recall is an important metric as it reflects the ability of the model to detect \textit{Crossing}; our model ranks second to Global PCPA~\cite{yang2022predicting} with $4\%$ reduction in recall and $99.99\%$ reduction in FLOPS and $99.97\%$ reduction in the number of trainable parameters.}

\begin{table}[] 
\caption{{JAADbeh: Comparative analysis of \textbf{(A)} 
baseline model (M2) and  CSE(M2+M3), \textbf{(B)} baseline model CSE(M2+M3) and additional model M1. The (baseline) models are shown between parenthesis.}}
\label{tab:jaadbeh_correlation_1}\resizebox{\columnwidth}{!}{
\begin{tabular}{|llll|}
\hline
\multicolumn{4}{|c|}{JAADbeh}                                                     \\ \hline
\multicolumn{2}{|c||}{(A)}  &    \multicolumn{2}{|c|}{(B)}                                               \\ \hline
\multicolumn{1}{|l|}{Baseline\textbackslash{}(Baseline)+additional}                                                         & \multicolumn{1}{l||}{(M2)+M3} & \multicolumn{1}{l|}{Baseline\textbackslash{}(Baseline)+additional}                                                              & \begin{tabular}[c]{@{}l@{}}(M2+M3)\\ +M1\end{tabular} \\ \hline
\multicolumn{1}{|l|}{(M2) false predictions}                                                                                 & \multicolumn{1}{l||}{0.4}     & \multicolumn{1}{l|}{(M2+M3) false predictions}                                                                                   & 0.35                                                  \\ \hline
\multicolumn{1}{|l|}{M3 false predictions}                                                                                   & \multicolumn{1}{l||}{0.39}    & \multicolumn{1}{l|}{M1 false predictions}                                                                                        & 0.42                                                  \\ \hline
\multicolumn{1}{|l|}{Common predictions}                                                                            & \multicolumn{1}{l||}{0.63}    & \multicolumn{1}{l|}{Common predictions}                                                                                 & 0.66                                                  \\ \hline
\multicolumn{1}{|l|}{Common false predictions}                                                                               & \multicolumn{1}{l||}{0.21}    & \multicolumn{1}{l|}{Common false predictions}                                                                                    & 0.21                                                  \\ \hline
\multicolumn{1}{|l|}{Actual false predictions}                                                                               & \multicolumn{1}{l||}{0.35}    & \multicolumn{1}{l|}{Actual false predictions}                                                                                    & 0.40                                                  \\ \hline
\multicolumn{1}{|l|}{Union of false predictions}                                                                                 & \multicolumn{1}{l||}{0.58}    & \multicolumn{1}{l|}{Union of false predictions}                                                                                      & 0.56                                                  \\ \hline
\multicolumn{1}{|l|}{\begin{tabular}[c]{@{}l@{}}Correlation coefficient \\ of M2 and M3 \\ confidence\end{tabular}} & \multicolumn{1}{l||}{0.41}    & \multicolumn{1}{l|}{\begin{tabular}[c]{@{}l@{}}Correlation coefficient \\ of (M2+M3) and M1 \\ confidence\end{tabular}} & 0.28                                                  \\ \hline
\end{tabular}}
\end{table}

\begin{table}[] 
\caption{{JAADbeh: Comparative analysis of \textbf{(A)} 
baseline model (M1) and  CSE(M1+M3), \textbf{(B)} baseline model CSE(M1+M3) and additional model M2. The (baseline) models are shown between parenthesis.}}
\label{tab:jaadall_correlation_1}
\resizebox{\columnwidth}{!}{
\begin{tabular}{|llll|}
\hline
\multicolumn{4}{|c|}{JAADall}                                                                                              \\ \hline
\multicolumn{2}{|c||}{(A)}  &    \multicolumn{2}{|c|}{(B)}                                               \\ \hline
\multicolumn{1}{|l|}{Baseline\textbackslash{}(Baseline)+additional}                                                         & \multicolumn{1}{l||}{(M1)+M3} & \multicolumn{1}{l|}{Baseline\textbackslash{}(Baseline)+additional}                                                              & \begin{tabular}[c]{@{}l@{}}(M1+M3)\\ +M2\end{tabular} \\ \hline
\multicolumn{1}{|l|}{(M1) false predictions}                                                                                          & \multicolumn{1}{l||}{0.19}    & \multicolumn{1}{l|}{(M1+M3)}                                                                                            & 0.17                                                  \\ \hline
\multicolumn{1}{|l|}{M3 false predictions}                                                                                            & \multicolumn{1}{l||}{0.32}    & \multicolumn{1}{l|}{M2}                                                                                                 & 0.36                                                  \\ \hline
\multicolumn{1}{|l|}{Common predictions}                                                                            & \multicolumn{1}{l||}{0.68}    & \multicolumn{1}{l|}{Common predictions}                                                                                 & 0.59                                                  \\ \hline
\multicolumn{1}{|l|}{Common false predictions}                                                                               & \multicolumn{1}{l||}{0.1}     & \multicolumn{1}{l|}{Common false predictions}                                                                                    & 0.06                                                  \\ \hline
\multicolumn{1}{|l|}{Actual false predictions}                                                                               & \multicolumn{1}{l||}{0.17}    & \multicolumn{1}{l|}{Actual false predictions}                                                                                    & 0.12                                                  \\ \hline
\multicolumn{1}{|l|}{Union false predictions}                                                                                 & \multicolumn{1}{l||}{0.41}    & \multicolumn{1}{l|}{Union false predictions}                                                                                      & 0.47                                                  \\ \hline
\multicolumn{1}{|l|}{\begin{tabular}[c]{@{}l@{}}Correlation coefficient \\ of M1 and M3 \\ confidence\end{tabular}} & \multicolumn{1}{l||}{0.284}   & \multicolumn{1}{l|}{\begin{tabular}[c]{@{}l@{}}Correlation coefficient \\ of (M1+M3) and M2 \\ confidence\end{tabular}} & 0.06                                                  \\ \hline
\end{tabular}}
\end{table}

\subsubsection{{Sensitivity} analysis}
{in this section, we investigate the causes for the differing CSE (M1+M2+M3) gain seen in Table~\ref{tab:AllData} between JAADbeh and JAADall. To this end, we study the false predictions of (A) the best performing single model in comparison with the best 2-model-based CSE approach (Table~\ref{tab:jaadbeh_correlation_1} Part (A) for JAADbeh and Table~\ref{tab:jaadall_correlation_1} Part (A) for JAADall) and (B) the best 2-model-based CSE approach in comparison with the 3-model-based CSE approach (Table~\ref{tab:jaadbeh_correlation_1} Part (B) for JAADbeh and Table~\ref{tab:jaadall_correlation_1} Part (B) JAADall).}
{As shown in Table~\ref{tab:AllData}, M2 has the best single-model performance (the baseline model) on JAADbeh whereas CSE(M2+M3) is the best 2-model-based CSE approach. As such, in Table~\ref{tab:jaadbeh_correlation_1}, M1 is the baseline model compared to CSE(M2+M3) (M3 is the additional model) in Part (A), whereas in Part (B), CSE(M2+M3) becomes the baseline compared with CSE(M1+M2+M3) (M1 is the additional model in this case). Similarly, M1 is the best single performing model thus, is considered the baseline for Part (A) of Table~\ref{tab:jaadall_correlation_1} with M3 as additional model; moreover, CSE(M1+M3) is baseline in Part (B) with M2 is additional.} 

{Ensemble learning has proved to be effective if training data and submodels are diverse, meaning model outputs are preferably independent or negatively correlated \cite{zhang2012ensemble}. Therefore, to examine the input (data) diversity of a merged CSE model, we first examine the output of the baseline and additional models and extract the distribution characteristics listed in Tables~\ref{tab:jaadbeh_correlation_1} and ~\ref{tab:jaadall_correlation_1}. Firstly, we calculate the ratio of ``Common Predictions" between each baseline and additional model which is defined as the number of samples in the test dataset assigned the same class by both. For instance, $63\%$ of the predicted classes ($\hat{y}$) using M1 (baseline) and M3 (additional) in Table~\ref{tab:jaadbeh_correlation_1} Part (A) are common to both models. Similarly, the ratio of ``Common False Predictions" indicates the percentage of the predicted samples that are commonly misclassified by both models; in Table~\ref{tab:jaadbeh_correlation_1} Part (A), this is $21\%$. Next, we calculate the ratio of ``Union of False Predictions" from the baseline and additional models which represents the cardinality of the union of misclassified samples from both models divided by the size of the test dataset; in Table~\ref{tab:jaadbeh_correlation_1} Part (A), this is $58\%$ of the test dataset. We then calculate the ``Actual False Predictions" for CSE models (Table~\ref{tab:jaadbeh_correlation_1} Part (A), this is $35\%$) and the Pearson product-moment correlation coefficients between the baseline and additional models (Table~\ref{tab:jaadbeh_correlation_1} Part (A), this is $41\%$). (Note: all scores are normalized by the number of samples in the test set, except for the correlation coefficient). The confusion matrices of the CSE models in Parts (A) and (B) of Tables~\ref{tab:jaadbeh_correlation_1} and ~\ref{tab:jaadall_correlation_1}} are shown in Figures \ref{fig:ConfBeh} and \ref{fig:ConfAll}, respectively, to facilitate the analysis.}

{In Table~\ref{tab:jaadbeh_correlation_1} Part (A), 
M2 and M3 have a high ratio of common false predictions, union of false predictions, and high correlation coefficients in their confidence outputs. In addition, CSE(M2+M3) and M1 also have high ratios in these three scores. In this case, the baseline model might not again much in performance when merged with an additional model using the CSE approach. On the other hand, in Table~\ref{tab:jaadall_correlation_1}, despite having non-negligible ratio of union of false predictions, M1 and M3 have a low common false predictions ratio and low correlation coefficient in their confidence outputs. The same applies to CSE(M1+M3) and M2 in Part(B). This means that the baseline and additional models benefit from their combination, making ensemble learning more effective, compared to Table\ref{tab:jaadbeh_correlation_1}.}

{Overall, in Table~\ref{tab:jaadbeh_correlation_1}, we observe a high correlation coefficient in confidence between the baseline and the additional models, compared to \ref{tab:jaadall_correlation_1}; in this case, a baseline model might not gain in performance from merging with an additional model. Moreover, we suspect that if both models have a high ratio of common false predictions, merged performance is likely to get worse compared to pre-merged performance. This can be seen by comparing the performance of M2 and CSE(M2+M3), as well as CSE(M2+M3) and M1 in Table~\ref{tab:jaadbeh_correlation_1}. In Table \ref{tab:jaadall_correlation_1}, however, JAADall has a lower correlation coefficient and common false predictions ratio even after the first merging. We conclude that it is the reason why the CSE(M1+M2+M3) results for JAADall reported in Table~\ref{tab:AllData} and confirmed by the confusion matrices in Figures~\ref{fig:ConfAll} are generally better than the results obtained with JAADbeh (see Figures~\ref{fig:ConfBeh}).}

\begin{figure} 
    \centering
  \subfloat[\footnotesize{CSE(M2+M3)}\label{1b}]{%
        \includegraphics[width=0.5\linewidth]{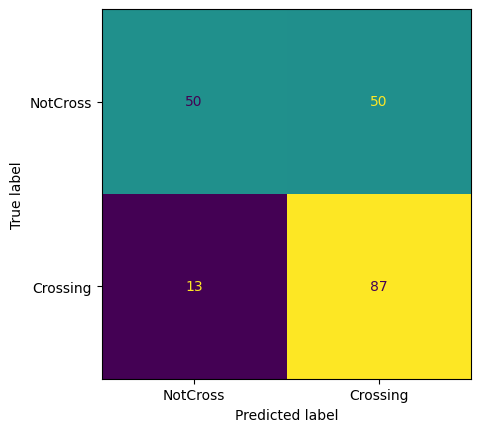}}
    \hfill
  \subfloat[\footnotesize{CSE(M1+M2+M3)}\label{fig:behm2}]{%
       \includegraphics[width=0.5\linewidth]{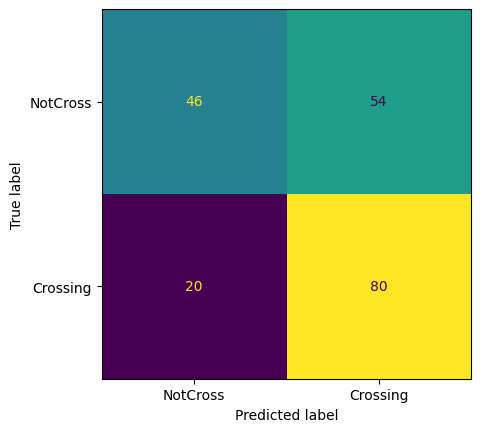}}
  \caption{Confusion matrices for CSE(M2+M3) and CSE(M1+M2+M3) in JAADbeh.}
  \label{fig:ConfBeh}
\end{figure}

\begin{figure}
    \centering
  \subfloat[\footnotesize{CSE(M1+M3)}\label{1b}]{%
        \includegraphics[width=0.5\linewidth]{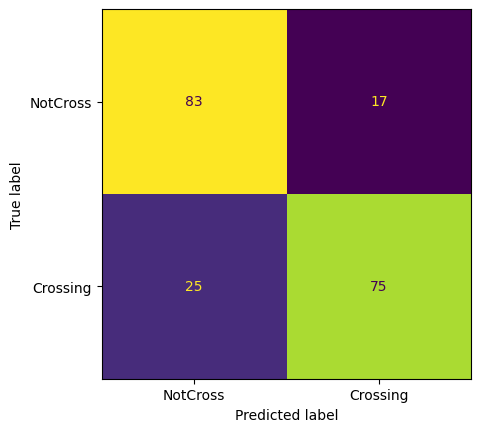}}
    \hfill
  \subfloat[\footnotesize{CSE(M1+M2+M3)}\label{fig:allm2}]{%
       \includegraphics[width=0.5\linewidth]{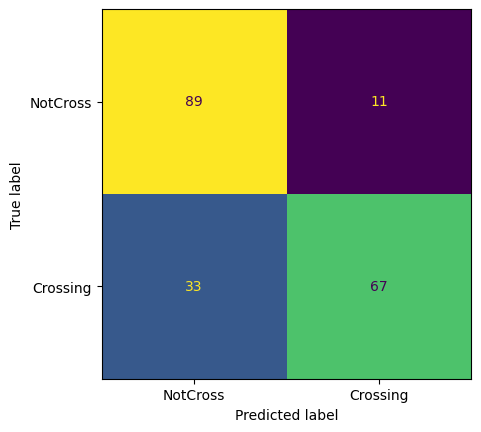}}
  \caption{Confusion matrices for CSE(M1+M3) and CSE(M1+M2+M3) in JAADall}
   \label{fig:ConfAll}
\end{figure}

\subsubsection{Generalization analysis}
{the proposed CSE model is built on a stacked ensemble approach that leverages multimodalities extracted from video footage. M2 learns the intent of pedestrians based on features that are particular to the JAAD dataset, i.e., Pedestrian Behavior, Traffic Background, and Vehicle Speed (see Figures~\ref{fig:behavior},~\ref{fig:traffic}, and~\ref{fig:vehspeed}). Due to the lack of such features in other PIP datasets, these cannot further validate the CSE(M1+M2+M3). On the other hand, M1 and M3 learn the intent of pedestrians based on the dynamics of pedestrian key-points during the time window and the relative location of the bounding box, respectively. It follows that both M1 and M3 can be readily applied to other PIP datasets to verify the generalize-ability of the proposed models. Indeed, a stacked ensemble model based on M1 and M3 was used for the IEEE Intelligent Transportation Systems Society student competition in Pedestrian Behavior in 2023, centered on the PSI dataset~\cite{chen2022psi} and secured the first prize. The obtained results using PSI and JAAD for training and the PSI test dataset determined by the competition are summarized in Table~\ref{tab:PSI}. As can be seen, the stacked ensemble learning model improves the performance of individual models M1 and M3, as for the JAADall dataset. These results demonstrate the generalization of the CSE approach, albeit without the M2 model.}

\begin{table*}[!h]
\caption{{Results of M1, M3, and CSE(M1+M3) models with PSI dataset (JAADbeh and JAADall results from Table~\ref{tab:AllData} repeated for ease of comparison)}}
\centering
\begin{tabular}{|l|lll|lll|lll|}
\hline
 & \multicolumn{3}{c|}{JAADbeh} & \multicolumn{3}{c|}{JAADall} & \multicolumn{3}{c|}{PSI} \\ \hline
 & \multicolumn{1}{l|}{ACC} & \multicolumn{1}{l|}{AUC} & F1-score & \multicolumn{1}{l|}{ACC} & \multicolumn{1}{l|}{AUC} & F1-score & \multicolumn{1}{l|}{ACC} & \multicolumn{1}{l|}{AUC} & F1-score \\ \hline
M1 & \multicolumn{1}{l|}{58\%} & \multicolumn{1}{l|}{65\%} & 58\% & \multicolumn{1}{l|}{81\%} & \multicolumn{1}{l|}{85\%} & 61\% & \multicolumn{1}{l|}{64\%} & \multicolumn{1}{l|}{71\%} & 61\% \\ \hline
M3 & \multicolumn{1}{l|}{\textbf{61\%}} & \multicolumn{1}{l|}{\textbf{69\%}} & \textbf{61\%} & \multicolumn{1}{l|}{68\%} & \multicolumn{1}{l|}{79\%} & 52\% & \multicolumn{1}{l|}{60\%} & \multicolumn{1}{l|}{65\%} & 57\% \\ \hline
CSE(M1+M3) & \multicolumn{1}{l|}{60\%} & \multicolumn{1}{l|}{70\%} & 59\% & \multicolumn{1}{l|}{\textbf{83\%}} & \multicolumn{1}{l|}{\textbf{87\%}} & \textbf{63\%} & \multicolumn{1}{l|}{\textbf{68\%}} & \multicolumn{1}{l|}{\textbf{72\%}} & \textbf{63\%} \\ \hline
\end{tabular}
\label{tab:PSI}
\end{table*}

\section{Conclusion}
This work examines the critical problem of pedestrian intent prediction (PIP), which raises safety concerns for vulnerable road users (VRU) in the advent of driver-assisted and driver-less vehicles. Ensuring the safety of VRUs when these types of vehicles are present is crucial, as it impacts the adoption of active travel in urban areas. An increase in the uptake of active travel accelerates the transition to sustainable mobility and transportation solutions.

With growing concerns about the carbon footprint of artificial intelligence, we propose a Contextual Stacked Ensemble-learning (CSE) approach to analyze visual data collected from ego-vehicles. Our method achieves a $99.97\%$ reduction in computational complexity compared to state-of-the-art techniques while demonstrating competitive performance with Pedgraph+ and Global PCPA. The effectiveness of our approach is built on three main pillars: the first focuses on data compression through skeleton-ization; the second emphasizes contextualization based on a thorough analysis of attributes inferred from visual data; and the third involves trajectory analysis of VRUs.

An extension of this work might examine how much information the network can contain and evaluate our ensemble technique presented in this paper. Moreover, we argued that when ensemble learning can(not) boost performance and how to evaluate input data and model diversity. Finally, the training of the three different output heads is done separately instead of simultaneously. Training in a multi-task formulation on multiple GPUs could be explored in the future.

\bibliographystyle{ieeetr}
\bibliography{sample}

\begin{IEEEbiography}[{\includegraphics[width=1in,height=1.25in,clip,keepaspectratio]{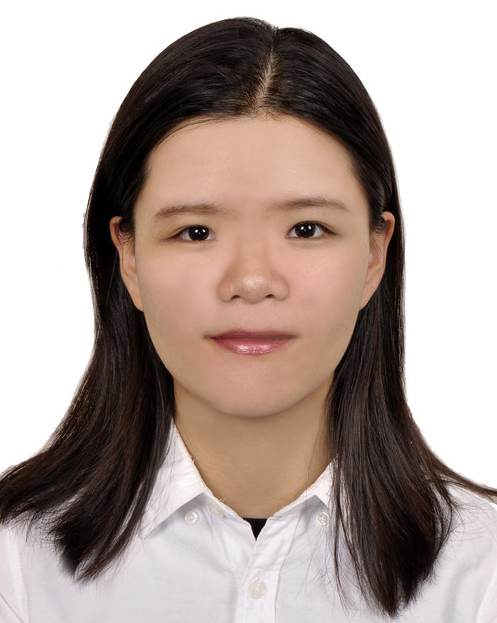}}]{Chia-Yen Chiang}(Student Member, IEEE) 
is a PhD student at Queen Mary University who is interested in AI applications for smart cities. Her interests involve building deep learning and machine learning models in multi-media resources such as video and audio. Previously, she worked on distributed acoustic sensing (DAS) time series data to detect vehicle size and its occupancy. She currently works on CCTV footage/GoPro videos for predicting traffic behaviors from vulnerable road users such as pedestrians and cyclists.
\end{IEEEbiography}

\begin{IEEEbiography}[{\includegraphics[width=1in,height=1.25in,clip,keepaspectratio]{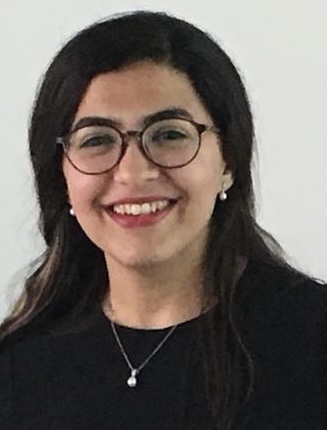}}]{Yasmin Fathy}
is a Research Associate at the Department of Engineering, University of Cambridge, and a Fellow of the Higher Education Academy. She is an active reviewer and a technical program committee member for multidisciplinary and international journals and conferences. She is also a member of the Cambridge Center for Data-Driven Discovery (C2D3). Her research interests include IoT, data analytics, digital twins, and applied machine learning.
\end{IEEEbiography}

\begin{IEEEbiography}[{\includegraphics[width=1in,height=1.25in,clip,keepaspectratio]{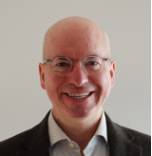}}]{Gregory Slabaugh}(Senior Member, IEEE) is currently Director of the Digital Environment Research Institute at Queen Mary University of London, and Professor of Computer Vision and AI with the School of Electronic Engineering and Computer Science. His research interests include computer vision, computational photography, and medical image computing.  He received a PhD in Electrical Engineering from Georgia Institute of Technology.  He has over 200 peer-reviewed publications and holds 38 granted patents.  His research has been funded by the EPSRC, BBSRC, MRC, Innovate UK and the European Commission. 
\end{IEEEbiography}

\begin{IEEEbiography}[{\includegraphics[width=1in,height=1.25in,clip,keepaspectratio]{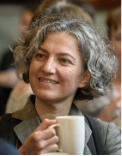}}]{Mona Jaber} (Senior Member, IEEE) is currently a Senior Lecturer on the Internet of Things (IoT) with the School of Electronic Engineering and Computer Science, Queen Mary University of London. Her research interests include zero-touch networks, the intersection of machine learning and IoT in the context of sustainable development goals, and IoT-driven digital twins. She secured the prestigious IEEE New Investigator Award for a project titled: “Distributed Acoustic Sensor system for Modelling Active Travel“ in which she investigates machine learning methods for processing optical fibre signals that represent active travel. She is the founder and director of the Digital Twins for Sustainable Development Goals (DT4SDG) research laboratory at QMUL. Mona is an executive committee of the IEEE United Kingdom and Ireland Woman in Engineering Affinity.\end{IEEEbiography}

\vfill

\end{document}